# A Coarse-to-Fine Approach for Urban Land Use Mapping Based on Multisource Geospatial Data

Qiaohua Zhou
*Department of Land Surveying and Geo-Informatics*
*The Hong Kong Polytechnic University*
Hong Kong, China
qiaohua.zhou@connect.polyu.hk

Rui Cao*
*Department of Land Surveying and Geo-Informatics*
*& Otto Poon C. F. Smart Cities Research Institute*
*The Hong Kong Polytechnic University*
Hong Kong, China
*Corresponding author, rucao@polyu.edu.hk

*Abstract*—Timely and accurate land use mapping is a long-standing problem, which is critical for effective land and space planning and management. Due to complex and mixed use, it is challenging for accurate land use mapping from widely-used remote sensing images (RSI) directly, especially for high-density cities. To address this issue, in this paper, we propose a coarse-to-fine machine learning-based approach for parcel-level urban land use mapping, integrating multisource geospatial data, including RSI, points-of-interest (POI), and areas-of-interest (AOI) data. Specifically, we first divide the city into built-up and non-built-up regions based on parcels generated from road networks. Then, we adopt different classification strategies for parcels in different regions, and finally combine the classified results into an integrated land use map. The results show that the proposed approach can significantly outperform baseline method that mixes built-up and non-built-up regions, with accuracy increase of 25% and 30% for level-1 and level-2 classification, respectively. In addition, we examine the rarely explored AOI data, which can further boost the level-1 and level-2 classification accuracy by 13% and 14%. These results demonstrate the effectiveness of the proposed approach and also indicate the usefulness of AOIs for land use mapping, which are valuable for further studies.

## I. Introduction

Timely and accurate monitoring and analysis of the spatiotemporal characteristics of major urban functional areas is of great significance for the coordinated development of land and space, especially in the current context of rapid urbanization across the world. The traditional method of producing large-scale land use maps has a heavy workload and long renewal cycle, which is difficult to meet the needs of practical applications [1, 2]. Remote sensing images (RSI) can capture rich spectral and texture physical features to facilitate efficient land use mapping [3-5], but they lack socio-economic features, which limits the land use mapping accuracy since land uses are closely related to socio-economic attributes. Advances in information and communication technologies (ICT) have greatly facilitated easy access to various urban big data, and thus provide the potential to fuse data from multiple sources for more effective urban land use mapping [6-8]. Among the emerging urban big data, point-of-interest (POI) and area-of-interest (AOI) data include rich spatial and attribute information of geographical entities that can reflect the socio-economic functional characteristics of urban built-up region, and can well complement the classification results of RSIs [9, 10].

However, RSIs, POIs, and AOIs have significantly different spatial coverage, distributions, and scales, which makes it difficult to fuse them effectively for land use mapping. To address this issue, in this paper, we propose a coarse-to-fine two-stage approach for urban land use mapping based on Random Forests (RF) models [11], integrating multisource geospatial data (including both RSIs, POIs and AOIs). The approach can realize rapid identification of large-scale urban functional areas, so as to provide reference for the coordinated development of land and space.

The major contributions of the paper are as follows. (1) We propose a two-stage approach for parcel-level urban land use classification, which can effectively exploit the characteristics of RSIs, POIs, and AOIs, and identify land use types in a progressive way from coarse- to fine-grained. (2) We further investigate into the use of AOIs, which have rarely been utilized, and demonstrate their usefulness in significantly improving the land use classification results.

The rest of the paper is organized as follows. Section 2 describes the study area and data. Section 3 elaborates the proposed coarse-to-fine approach and the evaluation method for parcel-level land use mapping. Section 4 reports and analyzes the results of experiments on Guangzhou, and further demonstrates the superiority of the proposed approach and also shows the effectiveness of AOIs. Finally, Section 5 concludes the paper with summary and discussion.

## II. Study Area and Dataset

### A. Study Area

Guangzhou, the capital of Guangdong Province, is located between 112 ° 57 'E to 114 ° 3'E and 22 ° 26 'N to 23 ° 56' N. The north east parts of the city are mountainous, whereas the remaining areas are plains, which is an integral part of the Pearl River Delta (Fig. 1). According to 2020 government data, Guangzhou has a total area of 7434.4 $km^2$ and a built-up region of 1350.95 $km^2$, with a complex and diverse urban structure, which poses great challenge for land use mapping.

### B. Data Collection

Road network and water surface vector data are from OpenStreetMap (OSM) (https://www.openstreetmap.org). According to the grade and size, the roads are divided into three levels. The first level includes motorway and trunk, the second level includes national and regional roads, and the third level is mainly local roads.

Sentinel-2 satellite images are obtained from Google Earth Engine(https://developers.google.com/earth-engine/datasets). In order to reduce the seasonal impact of vegetation and crops, the images in 2021 are divided by season. Through comparison, the cloud amount in spring is the least. Therefore,

This work was supported in part by the National Natural Science Foundation of China under Grant 42101472 and the Hong Kong Polytechnic University Start-Up under Grant BD41.

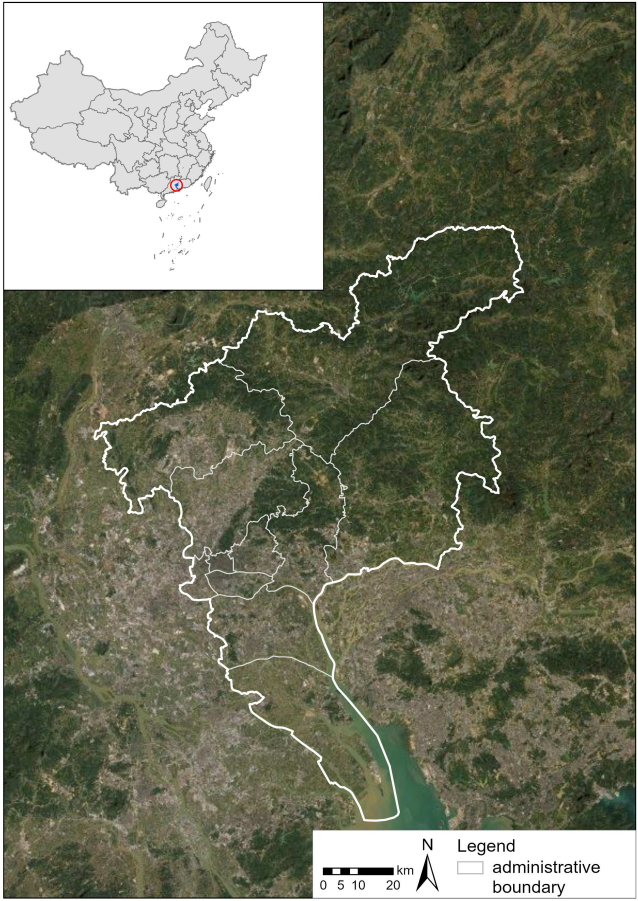

Fig. 1. Map of the study area of Guangzhou.

the images in spring (January to March) are selected for composition to remove cloud, and finally a cloudfree image of Guangzhou is obtained.

POI and AOI data are crawled from Baidu map, including attribute information such as name, category and geographic coordinates. The original data category is complex and there is data redundancy. Considering the large majority of POIs and AOIs are distributed in the build-up region, they are reclassified into 4 level-1 and 9 level-2 categories of the build-up region, and the data that are not related to land uses are removed. Finally, 367,732 POIs and 19,848 AOIs are obtained.

## III. METHODOLOGY

The flowchart of the research method is shown in Fig. 2, which includes four major procedures. (1) Classification scheme design and parcel generation; (2) Coarse-grained classification; (3) Fine-grained classification; (4) Accuracy assessment.

### A. Designing Classification Shcemes

Urban built-up regions generally refer to places dominated by artificial buildings and structures, while non-built-up regions are mainly composed of forest land, cultivated land and water areas. Based on the characteristics of built-up and non-built-up regions, the standard for urban land classification, and the actual situation of Guangzhou, the classification scheme is designed to be 8 level-1 and 16 level-2 categories (including categories of both built-up and non-built-up regions), as shown in TABLE I.

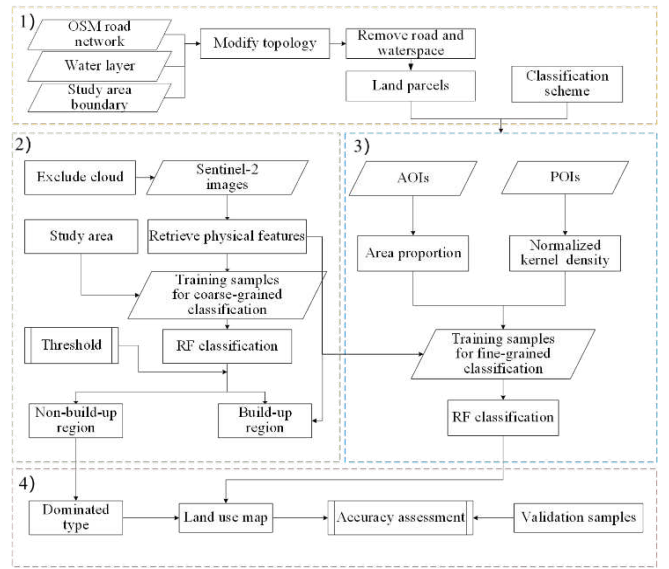

Fig. 2. Flowchart of the research method including four major procedures. 1) Classification scheme design and parcel generation; 2) Coarse-grained classification; 3) Fine-grained classification; 4) Accuracy assessment.

### B. Generating Land Parcels Based on Road Networks

This study uses land parcels as the basic unit, and assumes that the same land parcel has homogeneous socio-economic function. Roads and water are the boundaries of natural division of urban areas; thus, we use administrative boundaries, OSM road networks, and water boundaries to divide urban land into parcels. In order to obtain more accurate parcel polygons, topology modification of OSM road networks is needed. For roads with suspension points, if the suspension points are less than 500m away from adjacent roads, they will be extended. If the road length is less than 500m, they will be trimmed. After the topology modification, referring to the Code for Design of Urban Road Engineering of China (CJJ37 - 2012) and the actual road conditions of Guangzhou, the primary, secondary and tertiary roads are widened by 40m, 20m and 10m respectively to generate road space. Finally, the road and water space are removed in the administrative division and 3,645 parcels are generated (Fig. 3).

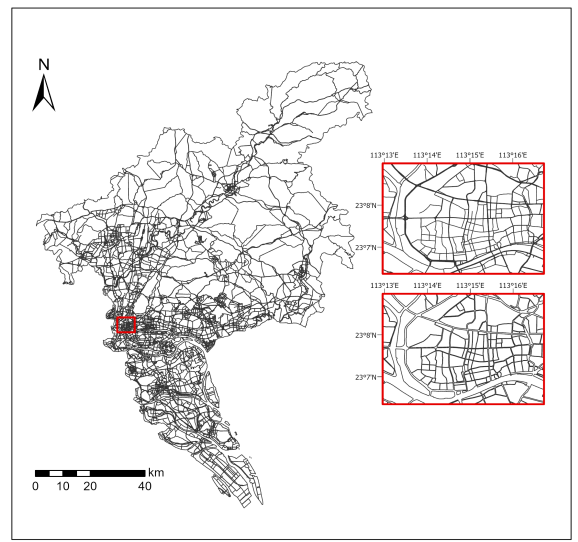

Fig. 3. Distribution of urban land use parcels generated from OSM road networks. (A) Overall pattern; (B) and (C) are zoomed-in views (red frame) of the original road networks and the segmented land parcels, respectively.

TABLE I. LAND USE CLASSIFICATION SCHEMES.

| Level-0 | Level-1 | Level-2 | Description |
|---|---|---|---|
| Non-built-up region (NBUR) | Agriculture (A) | Cropland (Cro) | Farm lands |
| | | Orchard (Orc) | Parcels planted with fruit trees |
| | | Aquaculture (Aqu) | Fish pond and aquaculture land |
| | Green Space (G) | Forest (For) | Trees with distinct canopy textures |
| | | Shrubland (Shr) | Shrubs with textures finer than trees |
| | Waterbody (W) | Waterbody (W) | Natural or artificial waterbodies |
| | Undeveloped (U) | Undeveloped (U) | Bare land or land under construction |
| Built-up region (BUR) | Residential (R) | Village (Vil) | Urban villages and rural areas |
| | | Community (Com) | Urban residential building groups |
| | Commercial (C) | Marketing (Mar) | Consumption and leisure land |
| | | Service building (Ser) | Office building for service industry |
| | Industrial (I) | Industrial (I) | Manufacturing, warehousing, mining |
| | Public Service (P) | Medical (Med) | Hospitals |
| | | Educational (Edu) | Education and scientific research institutions |
| | | Government (Gov) | Government and social organizations |
| | | Transportation (Tra) | Aviation, railway, passenger station and port land |

## C. Coarse-grained Classification

### 1) Extracting physical features from RSI

To better distinguish feature information, we first retrieved Normalized Difference Vegetation Index (NDVI) and Normalized Difference Water Index (NDWI) from the Sentinel-2 imagery.

$$NDVI = (NIR - RED)/(NIR + RED) \quad (1)$$
$$NDWI = (GREEN - NIR)/(GREEN + NIR) \quad (2)$$

where, NIR is band 8 of sentinel-2, RED is band 4, and GREEN is band 3.

Then, the mean values of RED, GREEN, BLUE (band2), NIR, SWIR1 (band11), NDVI and NDWI were extracted as the physical features for classification.

### 2) Annotating samples

By referring to the Sentinel-2 image (including both true-color and false-color composite images) and high-resolution images from Google Earth and Baidu map, we annotate the training samples with prominent characteristics for both coarse- and fine-grained classification. For example, the texture of forest land is rough, while the texture of cultivated land is fine and the shape is regular; the density of villages is usually high; a community is generally a regularly arranged building group including some green vegetation; and the industrial land is usually large in scale but low in intensity. Based on these characteristics, we collected 127 samples for coarse-grained classification and 124 for fine-grained classification. Some typical examples of different land use types are shown in Fig. 4. The numbers of training samples for different levels of classification are presented in TABLE II.

### 3) Identification of built-up and non-built-up regions

Based on the coarse-grained training samples, RF model can be trained and then leveraged to classify the Sentinel-2 image of the study area in pixel level. According to the classification results, the area proportion of built-up region in each parcel is calculated. Referring to the previous research [12, 13] and combined with the actual situation of Guangzhou, the threshold for distinguishing built-up region from non-built-up region is set to 0.37.

TABLE II. OVERVIEW OF THE COLLECTED TRAINING SAMPLES[a].
(COARSE: COARSE-GRAINED CLASSIFICATION; FINE: FINE-GRAINED CLASSIFICATION)

| Level-1 | Num. (coarse) | Num. (fine) | Level-2 | Num. (coarse) | Num. (fine) |
|---|---|---|---|---|---|
| Agriculture | 24 | 17 | Cropland | 7 | 6 |
| | | | Orchard | 9 | 5 |
| | | | Aquaculture | 8 | 6 |
| Green Space | 21 | 14 | Forest | 14 | 7 |
| | | | Shrubland | 7 | 7 |
| Waterbody[b] | 12 | - | Waterbody[b] | 12 | - |
| Undeveloped | 8 | 2 | Undeveloped | 8 | 2 |
| Residential | 14 | 25 | Village | 8 | 12 |
| | | | Community | 6 | 13 |
| Commercial | 13 | 24 | Marketing | 5 | 10 |
| | | | Service building | 8 | 14 |
| Industrial | 11 | 12 | Industrial | 11 | 12 |
| Public Service | 24 | 30 | Medical | 6 | 6 |
| | | | Educational | 5 | 6 |
| | | | Government | 8 | 9 |
| | | | Transportation | 5 | 9 |

a. A sample for coarse-grained classification is a block of pixels of the same category, while a sample for fine-grained classification is a land parcel of a specific category.
b. The category of *waterbody* is not applicable for fine-grained classification as there are no parcels that meet the criterion.

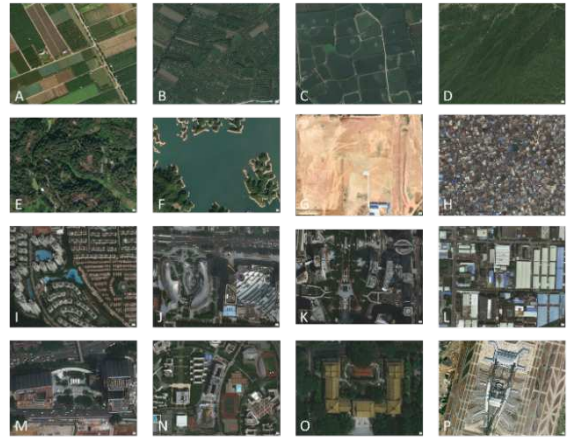

Fig. 4. Typical examples of training samples of different land use types. (A) Cropland; (B) Orchard; (C) Aquaculture; (D) Forest; (E) Shrubland; (F) Waterbody; (G) Undeveloped; (H) Village; (I) Community; (J) Marketing; (K) Service building; (L) Industrial; (M) Medical; (N) Educational; (O) Government; (P) Transportation.

If the proportion of built-up region of a parcel is greater than the threshold, the parcel is identified as built-up region, otherwise it is non-built-up region.

*D. Fine-grained Classification*

  *1) Extracting socio-economic features from POIs and AOIs*

POIs are discrete points. In order to densify the influence of POIs, we calculate the kernel density of different POI categories, and then normalize them to eliminate the volume differences of different categories to make them comparable [14].

$$K_{norm} = \frac{K - K_{min}}{K_{max} - K_{min}} \quad (3)$$

where, $K_{norm}$ is the normalized value of the POI kernel density map, and $K_{min}$ and $K_{max}$ are the minimum and maximum values in the POI density maps, respectively.

For AOIs, the area proportion of each category in a parcel is leveraged as an input feature for the RF model.

$$P_i = \frac{\sum_1^n A_i}{A_{parcel}} \quad (4)$$

where, $P_i$ is the area proportion of category *i*, $A_i$ is the area of each AOI of category *i* within a parcel, *n* is the number of AOIs of category *i* within a parcel, and $A_{parcel}$ is the area of this parcel.

  *2) Classification in non-built-up region*

For non-built-up region, the classification result of pixel scale at the coarse classification stage is used. By summarizing the area of each category, the area proportion of each category is used to determine the function of the parcel, the one with the largest proportion will be identified as the land use category for the parcel.

  *3) Classification in built-up region*

For built-up region, another RF model can be trained based on the annotated training samples of fine-grained labels. Then, the trained model is used to recognize the functions of land parcels [15]. Specifically, for each parcel, the input features include spectral characteristics of RSIs, normalized kernel density map of all categories, and the area proportions of all categories of AOIs.

*E. Accuracy Assessment*

We use the confusion matrix to evaluate the accuracy of land use classification. 255 validation parcels are generated in the study area by stratified sampling, including 150 built-up parcels and 105 non-built-up parcels. The land use types of the parcels are determined by visually inspecting Google Earth satellite imagery and Baidu map by local experts. Finally, the confusion matrices of level-1 and level-2 are established. In addition, metrics including overall accuracy (OA), Kappa coefficient, user accuracy (UA), producer accuracy (PA) are also used for classification performance evaluation.

## IV. RESULTS AND ANALYSIS

*A. Overall Results*

  *1) Coarse-grained land use mapping results*

The results of dividing built-up and non-built-up regions are shown in Fig. 5 and TABLE III. It can be seen that the spatial division of these two regions is in line with the actual situation to a great extent. The OA is 94% and the kappa coefficient is 0.87, which demonstrates the effectiveness of using physical features extracted from RSIs to separate built-up and non-built-up regions.

The PA of the built-up region is 100%, which means all parcels with the ground truth of built-up region were successfully identified. The PA of the non-built-up region is 89%, where parcels misclassified mainly include both built-up region and non-built-up region, such as villages and farmland, residential communities and green space. There are mainly two possible sources of errors. One is the error caused by spectral feature classification at the pixel scale. The other is the error caused by the classification based on the area proportion threshold of built-up and non-built-up region at the parcel scale. When the area proportion of the two is close to the threshold, it is difficult to decide the category.

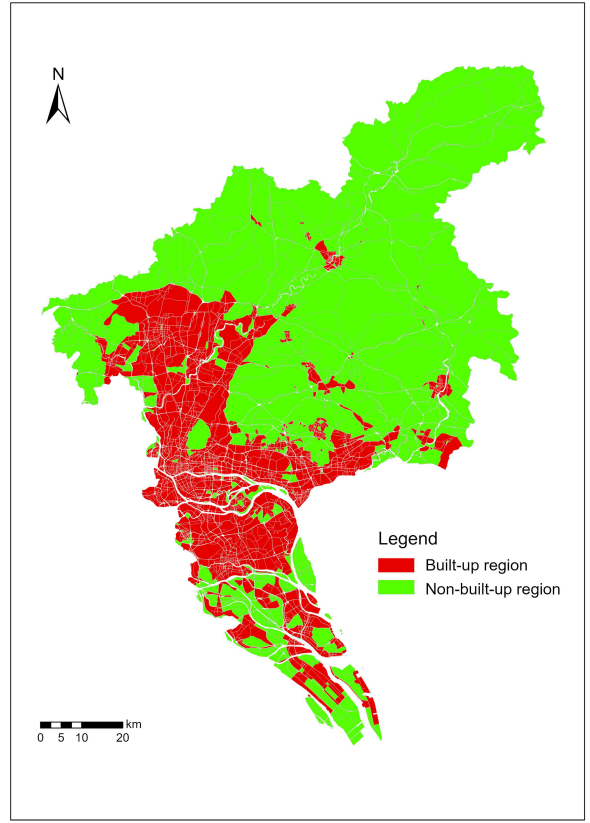

Fig. 5. Coarse-grained land use mapping results.

TABLE III. CONFUSION MATRIX OF COARSE-GRAINED CLASSIFICATION.

| Level-0 | NBUR | BUR | Total | UA | Kappa |
|---|---|---|---|---|---|
| Non-build-up region (NBUR) | **25** | 0 | 25 | 100% | 0 |
| Build-up region (BUR) | 3 | **20** | 23 | 87% | 0 |
| Total | 28 | 20 | **48** | 0 | 0 |
| PA | 89% | 100% | 0 | **94%** | 0 |
| Kappa | 0 | 0 | 0 | 0 | **0.87** |

  *2) Fine-grained land use mapping results*

The fine-grained classification maps are shown in Fig. 6(b) (level-1) and Fig. 6(e) (level-2). Based on features extracted from RSI and POIs, the two-stage approach can achieve an OA and kappa coefficient of 73% and 0.68 for leve-1 classification, while that of level-2 are 62% and 0.58, as shown in Fig. 7(b)(e) and Fig. 8(b)(e). We can see that the spatial distribution of land use categories is roughly consistent with the actual situation in Guangzhou, with green space distributed in the north, agriculture in the east, west and south, and the core built-up region of residential, commercial and public service land, and periphery the industrial land.

For the level-1 classification results, the PA and UA of undeveloped land, agriculture land and green space are all larger than 90%, which shows that the model performs well in the non-build-up region. For build-up region, residential and commercial land have the lowest PA (both 45%), while commercial and industrial land have the relatively low UA (56% and 50% respectively). Combined with the confusion matrix of level-2 classification, it can be seen that the misclassified residential land is mainly villages, most of which are misclassified as industrial land, and the misclassified commercial land is mainly the market land, which is mainly misclassified as communities. This is reasonable because in built-up region, factories are usually located together with urban villages, and residential communities are usually accompanied by consumption and entertainment facilities.

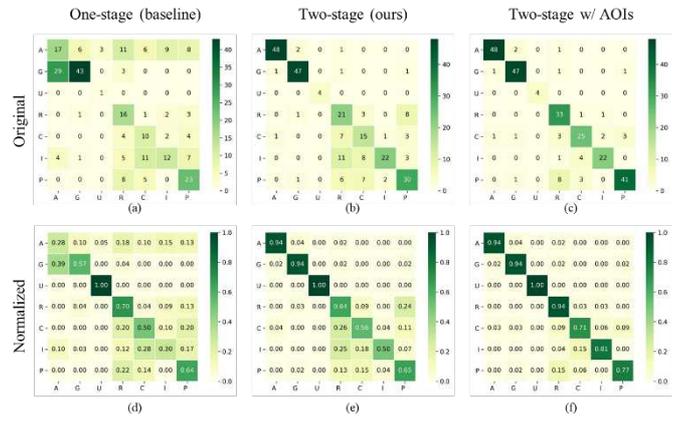

Fig. 7. Confusion matrices of level-1 classification. (a) One-stage approach (baseline), (b) Two-stage approach (ours), and (c) Two-stage approach with AOIs. The upper row (a,b,c) presents the original confusion matrices, while the bottom row (d,e,f) shows corresponding normalized confusion matrices.

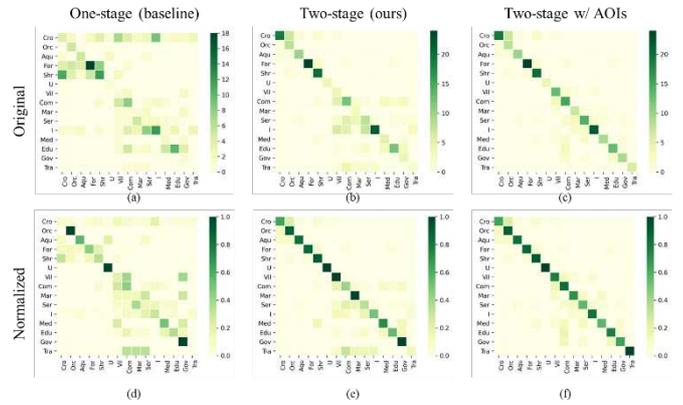

Fig. 8. Confusion matrices of level-2 classification. (a) One-stage approach (baseline), (b) Two-stage approach (ours), and (c) Two-stage approach with AOIs. The upper row (a,b,c) presents the original confusion matrices, while the bottom row (d,e,f) shows corresponding normalized confusion matrices.

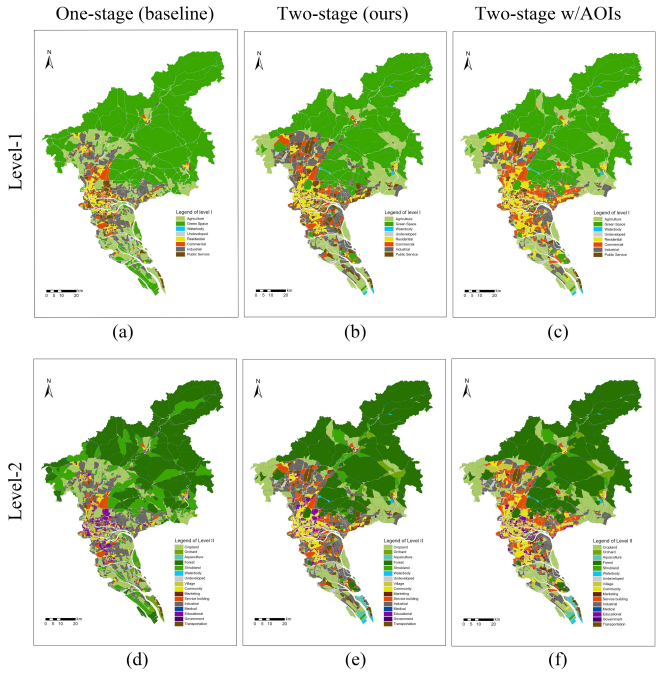

Fig. 6. Fine-grained land use mapping results of level-1 and level-2. (a) and (d) One-stage approach (baseline), (b) and (e) Two-stage approach (ours), (c) and (f) Two-stage approach with extra AOI data.

### B. Comparison between One- and Two-stage Approaches

The OA obtained by the baseline one-stage method (directly identifying the land use categories of the whole city in parcel level based on RSI and POI features) is relatively low, with level-1 accuracy of 48%, and level-2 accuracy of 32%, and the kappa coefficient of them are 0.37 and 0.26, respectively (Fig. 7(a)(d) and Fig. 8(a)(d)).

From the perspective of distinguishing non-built-up and built-up region, this method performed worse for the identification of built-up region, in which 37 of the 155 validation samples of built-up region were identified as non-built-up region (Fig. 6(a)(d)). After adopting the proposed two-stage approach, this number was reduced significantly to 3. From the internal subclassified results of non-built-up and built-up region, their average PA is 48% and 41% respectively, and have improved to 96% and 61% respectively after adopting the two-stage scheme. It can be seen that when RSI and POI are applied to non-built-up and built-up region for RF classification at the same time, whether it is to distinguish non-built-up and built-up region, or to finer categories, the results cannot compete with applying them to different regions separately. This may be because the vast majority of POI data are located in built-up region, and the model performance will be significantly impacted by the extremely uneven data distribution.

*C. Analysis of the Effects of AOIs*

At present, research on urban land identification involving AOIs is still relatively rare. Therefore, we set up a comparative experiment to compare the classification performance before and after adding AOIs as an extra data source.

Together with POI and spectral features, extra AOI features help improve the accuracy of level-1 classification to 86%, with a kappa coefficient of 0.83, and the OA and kappa coefficients of level-2 are 76% and 0.74 respectively (Fig. 7(c)(f) and Fig. 8(c)(f)). It can be seen that the results of both levels have been significantly improved. The classification maps of level-1 and level-2 are shown as Fig. 6(c) (f). It can be seen that compared with the results before adding AOI, the overall land pattern is more consistent with the actual situation. The residential, commercial and industrial land presented multiple cores rather than simple aggregation.

In the level-1 result, the PA an UA of all categories have been improved or remained unchanged. Among which, the significant increase of PA are in the categories of commercial (31%), residential (25%) and public service (24%). Combined with level-2 results, the PA of transportation land has improved most significantly from 33% to 100%. This may be because the traffic hub usually has a boundary of large floor area in AOI data, so it is easy to be identified compared with only one or several points in POI data, which is easy to be affected by the surrounding POIs. Similarly, since most villages, industrial parks, business centers and residential communities in AOI data have clear boundaries, the confusion between villages and industrial land, and between marketing and residential land have been alleviated to varying degrees. After adding AOI data, the PA of villages and marketing land improved by 45% and 32% respectively. At the same time, the importance analysis of features also shows that AOIs has the greatest contribution of the model, showing that AOI data can be used as an effective data source for urban land classification.

*D. Analysis of the Land Use Pattern of Guangzhou*

According to the parcel-level land use mapping results, the built-up region of Guangzhou is 1,868.48km$^2$, accounting for 25.1% of the whole city area. The proportions of residential land, commercial land and industrial land are similar in the built-up region, which account for 31.4%, 28.3% and 29.4% respectively, and remaining public service land hold a proportion of 10.9%.

Due to the influence of terrain, the land use pattern of Guangzhou presents a north-south spatial structure. The ecological space is mainly distributed in the north, while the agricultural space lays in the east and west wings and the southwest, and the urban space is relatively concentrated in the middle. Among them, the urban space roughly presents the concentric circle development model. The residential and public service land are mainly distributed in the urban core area, with commercial land closely distributed in the periphery, and the industrial land is concentrated in the east, south and north wing in a cluster development mode.

## V. Discussion and Conclusions

Automatic urban land use mapping is essential for timely and accurate urban land monitoring and management. RSIs are widely used, and can efficiently capture the physical elements and have high accuracy in the recognition of ground features in non-built-up region. However, it is often not sufficient to distinguish complex land use types in built-up region using RSIs alone. The emerging open social data include rich socio-economic information that can be leveraged to identify the functions of built-up region, but there is almost no data in non-built-up region. Due to their significant differences in spatial coverage, distributions, and scales, it is non-trivial to fuse them effectively for land use mapping.

To address this issue, in this paper, we propose a coarse-to-fine machine learning-based approach for urban land use mapping, integrating multisource RSI, POI and AOI data. Specifically, we first divide the city area into built-up region and non-built-up region, and then adopt different classification strategies according to their different characteristics. In addition, the rarely explored AOI data are added as an extra data source. AOIs focus on features with a certain scale, which can more directly reflect land use than POIs. The experimental results show that using the proposed coarse-to-fine classification strategy and combined with AOI data can produce a more detailed and accurate land use map at parcel level.

Our approach has achieved good results in generating detailed urban land use map on a large district, however, there are still some limitations. First, the quality of road networks directly determines the number and size of land parcels. The parcels formed by sparse road networks contain more mixed land use types, and therefore it will be difficult to identify the functions. Despite the easy access, the integrity and correctness of OSM road networks need to be improved. Future research can consider combining road networks and the segmentation from RSI to generate more detailed and consistent parcels. Second, the land use mapping results can be further enhanced by using more carefully selected features or integrating more sources of data. In this study, RSI, POI and AOI data are used. With the increasing availability of various urban big data, it is beneficial to include other data sources to further improve the classification results, such as human mobility data, which are complementary and can provide more comprehensive information for land use recognition.

## References

[1] J.-P. Donnay, and D. Unwin, "Modelling geographical distributions in urban areas," Remote Sensing and Urban Analysis, GISDATA 9, pp. 205-224, 2001.

[2] S.-S. Wu, X. Qiu, E. L. Usery, and L. Wang, "Using geometrical, textural, and contextual information of land parcels for classification of detailed urban land use," Annals of the Association of American Geographers, vol. 99, no. 1, pp. 76-98, 2009.

[3] D. Lu, and Q. Weng, "Use of impervious surface in urban land-use classification," Remote sensing of environment, vol. 102, no. 1-2, pp. 146-160, 2006.

[4] Huang, B., Zhao, B., & Song, Y, Urban land-use mapping using a deep convolutional neural network with high spatial resolution multispectral remote sensing imagery. Remote Sensing of Environment, 214, 73-86, 2018.

[5] M. Shaban, and O. Dikshit, "Improvement of classification in urban areas by the use of textural features: the case study of Lucknow city, Uttar Pradesh," International Journal of remote sensing, vol. 22, no. 4, pp. 565-593, 2001.

[6] R. Cao, J. Zhu, W. Tu, Q. Li, J. Cao, B. Liu, Q. Zhang, and G. Qiu, "Integrating aerial and street view images for urban land use classification," Remote Sensing, vol. 10, no. 10, pp. 1553, 2018.

[7] R. Cao, W. Tu, C. Yang, Q. Li, J. Liu, J. Zhu, Q. Zhang, Q. Li, and G. Qiu, "Deep learning-based remote and social sensing data fusion for urban region function recognition," ISPRS Journal of Photogrammetry and Remote Sensing, vol. 163, pp. 82-97, 2020.

[8] D. Chen, W. Tu, R. Cao, T. Shi, Y. Zhang, B. He, C. Wang, Qingquan. Li, "A hierarchical approach for fine-grained urban villages recognition fusing remote and social sensing data," *International Journal of Applied Earth Observation and Geoinformation*, vol. 106, p. 102661, 2022.

[9] G. McKenzie, K. Janowicz, and B. Adams, "A weighted multi-attribute method for matching user-generated points of interest," Cartography and Geographic Information Science, vol. 41, no. 2, pp. 125-137, 2014.

[10] S. Elwood, M. F. Goodchild, and D. Z. Sui, "Researching volunteered geographic information: Spatial data, geographic research, and new social practice," Annals of the association of American geographers, vol. 102, no. 3, pp. 571-590, 2012.

[11] L. Breiman, "Random forests," Machine learning, vol. 45, no. 1, pp. 5-32, 2001.

[12] G. Xian, and M. Crane, "Assessments of urban growth in the Tampa Bay watershed using remote sensing data," Remote sensing of environment, vol. 97, no. 2, pp. 203-215, 2005.

[13] F. Aguilera, L. M. Valenzuela, and A. Botequilha-Leitão, "Landscape metrics in the analysis of urban land use patterns: A case study in a Spanish metropolitan area," Landscape and Urban Planning, vol. 99, no. 3-4, pp. 226-238, 2011.

[14] T. Hu, J. Yang, X. Li, and P. Gong, "Mapping urban land use by using landsat images and open social data," Remote Sensing, vol. 8, no. 2, pp. 151, 2016.

[15] P. Gong, B. Chen, X. Li, H. Liu, J. Wang, Y. Bai, J. Chen, X. Chen, L. Fang, and S. Feng, "Mapping essential urban land use categories in China (EULUC-China): Preliminary results for 2018," 2020.